\pdfoutput=1
\documentclass[11pt]{article}

\usepackage[table]{xcolor}
\usepackage[final]{acl}

\usepackage{times}
\usepackage{latexsym}

\usepackage[T1]{fontenc}
\usepackage[utf8]{inputenc}

\usepackage{microtype}

\usepackage{inconsolata}

\usepackage{graphicx}

\usepackage{contour}

\newcommand{\hide}[1]{}

\newcommand{\AMADDA}{{\={A}}}
\newcommand{\AHAMZAUP}{{\^{A}}}
\newcommand{\WHAMZA}{{\^{w}}}
\newcommand{\AHAMZADN}{{\v{A}}}
\newcommand{\YHAMZA}{{\^{y}}}
\newcommand{\TAMARBUTA}{{$\hbar$}}
\newcommand{\TAMAR}{{$\hbar$}}
\newcommand{\THA}{{$\theta$}}
\newcommand{\DHA}{{\dh}}
\newcommand{\SHIN}{{\v{s}}}
\newcommand{\DAD}{{\v{D}}} %Z
\newcommand{\ZA}{{\v{D}}} %Z
\newcommand{\AYN}{{$\varsigma$}}
\newcommand{\GAYN}{{$\gamma$}}
\newcommand{\AMAQSURA}{{\'{y}}}
\newcommand{\AMAQ}{{\'{y}}}

\usepackage{arabtex}  
\usepackage{hebtex}  
\usepackage{textgreek}
\usepackage[utf8]{inputenc} % Support UTF-8 input
\DeclareUnicodeCharacter{0127}{\hbar}

\usepackage{booktabs}

\usepackage{utf8}
\usepackage{epstopdf}

\title{A Tale of Two Scripts:\\
Transliteration and Post-Correction for Judeo-Arabic}
\author{Juan Moreno Gonzalez,\textsuperscript{1} Bashar Alhafni,\textsuperscript{2} Nizar Habash\textsuperscript{3}\\
  \textsuperscript{1}University of Cambridge\\
  \textsuperscript{2}Mohamed bin Zayed University of Artificial Intelligence\\
  \textsuperscript{3}New York University Abu Dhabi\\
  \texttt{jm2553@cam.ac.uk}, \texttt{bashar.alhafni@mbzuai.ac.ae}, \texttt{nizar.habash@nyu.edu}
  }

\begin{document}
\maketitle

\setcode{utf8}
\setarab
\vocalize

%\textbf{\textsc{Hello}}
%\textsc{Hello} 

\begin{abstract}
Judeo-Arabic refers to Arabic variants historically spoken by Jewish communities across the Arab world, primarily during the Middle Ages. Unlike standard Arabic, it is written in Hebrew script by Jewish writers and for Jewish audiences. Transliterating Judeo-Arabic into Arabic script is challenging due to ambiguous letter mappings, inconsistent orthographic conventions, and frequent code-switching into Hebrew. In this paper, we introduce a two-step approach to automatically transliterate Judeo-Arabic into Arabic script: simple character-level mapping followed by post-correction to address grammatical and orthographic errors. We also present the first benchmark evaluation of LLMs on this task. Finally, we show that transliteration enables Arabic NLP tools to perform morphosyntactic tagging and machine translation, which would have not been feasible on the original texts. We make our code and data publicly available.\footnote{\url{https://github.com/CAMeL-Lab/jawhar}}
% nd frequent code-switching into Hebrew and Aramaic with partial morphological integration.
% Finally, we demonstrate the utility of our approach by enabling standard Arabic NLP tools to perform morphosyntactic tagging and machine translation into English on the transliterated text, which would not be feasible on the original Judeo-Arabic texts.
\end{list}
\end{abstract}

\section{Introduction}

Judeo-Arabic (JA) refers to Arabic varieties historically used by Jewish communities across the Arab world, primarily in the Middle Ages. Although closely related to regional Arabic dialects, JA is written in Hebrew script and incorporates elements from Hebrew. Thousands of JA texts are now available online, covering genres such as philosophy, biblical commentary, and Bible translations \cite{friedberg1999geniza,rustow2022geniza}. However, because JA texts were intended for readers familiar with Hebrew script, they remain largely inaccessible to Arabic speakers and incompatible with modern Arabic NLP tools, which typically assume Arabic script input.

\renewcommand{\arraystretch}{0.95}
\begin{table}[t]
\setlength{\tabcolsep}{4pt}
\centering
\small

\resizebox{0.49\textwidth}{!}{
\begin{tabular}{ccccl}
\toprule
\textbf{JA} & \textbf{Arabic} & \textbf{Dotless T} & \textbf{Dotted T} & \textbf{English} \\
\midrule

\RL{\sethebrew{קאל}} &
\begin{tabular}[c]{@{}c@{}}\RL{\setarab{قال}} \\ \textit{qAl}\end{tabular} &
\cellcolor{green!5}\begin{tabular}[c]{@{}c@{}}\RL{\setarab{قال}} \\ \textit{qAl}\end{tabular} &
\cellcolor{green!5}\begin{tabular}[c]{@{}c@{}}\RL{\setarab{قال}} \\ \textit{qAl}\end{tabular} &
\textit{said} \\

\RL{\sethebrew{אלכׄזרי}} &
\begin{tabular}[c]{@{}c@{}}\RL{\setarab{الخزري}} \\ \textit{Alxzry}\end{tabular} &
\cellcolor{red!5}\begin{tabular}[c]{@{}c@{}}\RL{\setarab{الكزري}} \\ \textit{Alkzry}\end{tabular} &
\cellcolor{green!5}\begin{tabular}[c]{@{}c@{}}\RL{\setarab{الخزري}} \\ \textit{Alxzry}\end{tabular} &
\textit{al-Khazari} \\

, &
: &
\cellcolor{yellow!15}, &
\cellcolor{yellow!15}, &
, \\

\RL{\sethebrew{וכיף}} &
\begin{tabular}[c]{@{}c@{}}\RL{\setarab{وكيف}} \\ \textit{wkyf}\end{tabular} &
\cellcolor{green!5}\begin{tabular}[c]{@{}c@{}}\RL{\setarab{وكيف}} \\ \textit{wkyf}\end{tabular} &
\cellcolor{green!5}\begin{tabular}[c]{@{}c@{}}\RL{\setarab{وكيف}} \\ \textit{wkyf}\end{tabular} &
\textit{and how} \\

\RL{\sethebrew{דׄלך}} &
\begin{tabular}[c]{@{}c@{}}\RL{\setarab{ذلك}} \\ \textit{{\THA}lk}\end{tabular} &
\cellcolor{red!5}\begin{tabular}[c]{@{}c@{}}\RL{\setarab{دلك}} \\ \textit{dlk}\end{tabular} &
\cellcolor{green!5}\begin{tabular}[c]{@{}c@{}}\RL{\setarab{ذلك}} \\ \textit{{\THA}lk}\end{tabular} &
\textit{is that} \\

\RL{\sethebrew{והברכות}} &
\begin{tabular}[c]{@{}c@{}}``\RL{\setarab{والتسبيحات}}'' \\ \textit{w``AltsbyHAt''}\end{tabular} &
\cellcolor{red!5}\begin{tabular}[c]{@{}c@{}}\RL{\setarab{والبركوت}} \\ \textit{wAlbrkwt}\end{tabular} &
\cellcolor{red!5}\begin{tabular}[c]{@{}c@{}}\RL{\setarab{والبركوت}} \\ \textit{wAlbrkwt}\end{tabular} &
\begin{tabular}[l]{@{}l@{}}\textit{and the} \\ \textit{blessings}\end{tabular}\\

\RL{\sethebrew{כלפה}} &
\begin{tabular}[c]{@{}c@{}}\RL{\setarab{كلفة}} \\ \textit{klf{\TAMARBUTA}}\end{tabular} &
\cellcolor{yellow!5}\begin{tabular}[c]{@{}c@{}}\RL{\setarab{كلفه}} \\ \textit{klfh}\end{tabular} &
\cellcolor{yellow!5}\begin{tabular}[c]{@{}c@{}}\RL{\setarab{كلفه}} \\ \textit{klfh}\end{tabular} &
\textit{burden} \\

\RL{\sethebrew{זאידה}} &
\begin{tabular}[c]{@{}c@{}}\RL{\setarab{زائدة}} \\ \textit{zA{\YHAMZA}d{\TAMARBUTA}}\end{tabular} &
\cellcolor{yellow!5}\begin{tabular}[c]{@{}c@{}}\RL{\setarab{زايده}} \\ \textit{zAydh}\end{tabular} &
\cellcolor{yellow!5}\begin{tabular}[c]{@{}c@{}}\RL{\setarab{زايده}} \\ \textit{zAydh}\end{tabular} &
\textit{extra} \\

. & . &
\cellcolor{green!5}. &
\cellcolor{green!5}. &
. \\

\bottomrule
\end{tabular}
}
\caption{A sentence in Judeo-Arabic (Hebrew script), aligned with Arabic script, Arabic transliterations from dotless (Dotless T) and dotted (Dotted T) Hebrew, and English glosses.  The sentence is `Al-Khazari said: And how is that when blessings are an extra burden?'}
\vspace{-20pt}
\label{tab:first-example}
\end{table}

\renewcommand{\arraystretch}{1}

Transliterating JA into Arabic script is therefore a crucial step for making these texts accessible to Arabic readers and for ensuring compatibility with modern Arabic NLP tools.
% As such, transliterating Judeo-Arabic from Hebrew script to Arabic script is a necessary step towards making these texts accessible to Arabic speakers and compatible with modern Arabic NLP tools. 
However, this task is challenging due to ambiguous character mappings between the scripts, inconsistent orthographic conventions, and frequent 
%borrowings from Hebrew and Aramaic (Table~\ref{tab:first-example}).
Hebrew borrowings (Table~\ref{tab:first-example}).
To address this, we \textbf{propose a two-step approach}: an initial transliteration stage using a rule-based method, followed by a post-correction step to resolve orthographic and grammatical errors. We also present \textbf{the first benchmark evaluation of large language models (LLMs)} on this task.
% and compare our results to the most recent prior work, establishing a clear basis for progress on JA transliteration.
Notably, \textbf{our approach requires no training on JA data}, yet outperforms previous work.
Finally, we demonstrate the \textbf{downstream impact of transliteration} by enabling morphosyntactic tagging and Arabic-to-English machine translation (MT) on JA texts.
{Our results demonstrate the viability of this approach: post-correction significantly improves transliteration quality, and the resulting outputs can be effectively processed by Arabic NLP tools for both morphosyntactic tagging and MT.}

\section{Background and Related Work}
\label{sec:related}

\subsection{Judeo-Arabic Linguistic Facts}
\label{sec:ling-facts}
Judeo-Arabic (JA) is the language of the Jews from the Arabo-Islamic world. Its history spans from pre-Islamic Arabia to the present day and went through several different periods \cite{blau1961grammar,blau1965emergence}.
%\footnote{For a comprehensive overview of Judeo-Arabic see \newcite{blau1965emergence} and \newcite{blau1961grammar}.} 
This paper focuses on Classical JA, which developed in parallel to Classical Arabic in the territories under Muslim rule around the Mediterranean and the Middle East during the Middle Ages. JA is a Jewish language. It was written by Jews and its intended audience were other Jews. This explains two of its most distinctive characteristics. First, JA is written in Hebrew characters. Second, it incorporates some vocabulary from Hebrew and, less commonly, Aramaic. One should be cautious not to describe JA just as Arabic written in Hebrew characters \cite{stillman2010judeo}. Most 
%of the 
Jews from the medieval Mediterranean and Middle East lived under Muslim rule and spoke Arabic as their mother tongue. However, the most common writing system they used was Hebrew, not Arabic. From a very young age, Jewish children were taught Hebrew for religious purposes. Some of them would not be able to read or write Arabic, despite reading and writing in JA. Thus, for them JA was not a direct transliteration of Arabic, but rather the obvious way of registering their language in writing. For this same reason, some scholars have approached JA as an Arabic dialect \cite{mansour1991jewishbaghdadi,Khan:1997,gebski2024grammar}.

Transliterating JA written in Hebrew to Arabic characters is as artificial as transliterating it to Latin script. Nevertheless, this approach is justified when the goal is to make JA accessible to Arabic readers who do not read Hebrew, or to enable the use of NLP tools designed for Arabic. 
%Because the idea of Arabic to Hebrew transliteration is not essential to Judeo-Arabic, trying to transliterate Judeo-Arabic from Hebrew to Arabic characters entails some challenges.
This transliteration task entails many challenges.

The Arabic \textit{abjad} has twenty-eight basic letters and some additional special characters (e.g., Hamzated Alifs and Teh Marbuta); the Hebrew \textit{abjad} has twenty-two. 
% There is no one-to-one mapping of Arabic and Hebrew characters. The Arabic letters that do not correspond to a sound represented in the Hebrew \textit{abjad} are represented by a Hebrew letter with an upper dot, in imitation of the Arabic alphabet (see Figure \ref{tab:map-characters} in Appendix \ref{app:char-map}). There are seven Hebrew letters that have an upper dot diacritic variant in Judeo-Arabic: 
% \sethebrew 
% %\<ג ,ד ,ה ,ט ,כ ,צ ,ת>.
% \<גדהטכצת> \textit{jdhTkSt}.
While many Hebrew letters have one-to-one or many-to-one mappings to Arabic letters \cite{blau1961grammar,judeoarabic2020orthographies}, some  Arabic letters without corresponding sounds in the Hebrew \textit{abjad} are represented by a Hebrew letter with \textit{an upper dot}, imitating the Arabic alphabet (see Table~\ref{tab:map-characters}).
Seven Judeo-Arabic Hebrew letters have an upper dot diacritic variant: 
\sethebrew  \<גדהטכצת> \textit{jdhTkSt}.\footnote{The upper dot is sometimes used for other purposes, such as indicating abbreviations (JA: \sethebrew\<וגׄ> \textit{w{\GAYN}},  Ar: \setarab\<وغير> \textit{w{\GAYN}yr}, `etc') or indicating that a letter has a numeric value (\sethebrew\<מׄדׄ>, ``44'').}\textsuperscript{,\ref{translit}}

Unfortunately, the use of diacritical marks or the Hebrew-Arabic character mapping has never been standardized. Thus, the rules followed by different texts may vary \cite{judeoarabic2020orthographies}. Furthermore, common errors of omission or misplacement of the upper dot can lead to erroneous conversions, \sethebrew\<בלגהׄ> \setarab\<بلجة> \textit{blj{\TAMARBUTA}}\footnote{\label{translit}Arabic 
 transliteration is presented in the %one-to-one
  Habash-Soudi-Buckwalter transliteration scheme \cite{Habash:2007:arabic-transliteration}: \\%(in alphabetical order)\\
  \setarab
  \addtolength{\tabcolsep}{-5.5pt}
\begin{tabular}{cccccccccccccccccccccccccccc}
<ا> & <ب> & <ت> & <ث> & <ج> & <ح> & <خ> & <د> & <ذ> & <ر> & <ز> & <س> & <ش> & <ص> & <ض> & <ط> & <ظ> & <ع> & <غ> & <ف> & <ق> & <ك> & <ل> & <م> & <ن> & <ه> & <و> & <ي> \\
{\AHAMZAUP} & b & t & {\THA} & j & H & x & d & {\DHA} & r & z & s & {\SHIN} & S & D & T & {\DAD} & {\AYN} & {\GAYN} & f & q & k & l & m & n & h & w & y\\
\end{tabular}
 and additionally: '~<ء>, {\AHAMZAUP}~<أ>,
  {\AHAMZADN}~<إ>, {\AMADDA}~<آ>, {\WHAMZA}~<ؤ>, {\YHAMZA}~<ئ>,
  {\TAMARBUTA}~<ة>, {\AMAQSURA}~<ى>. }
instead of the correct \sethebrew\<בלגׄהׄ> \setarab\<بلغة> \textit{bl{\GAYN}{\TAMARBUTA}} `in a language', or \sethebrew\<והׄדא> \setarab\<وةدا> \textit{w{\TAMARBUTA}dA} instead of the correct \sethebrew\<והדׄא> \setarab\<وهذا> \textit{wh{\DHA}A} `and this'.
Furthermore, not every Arabic character is explicitly marked in JA. The Hamza at the end of the word is often dropped (\sethebrew\<גא> \setarab\<جا> \textit{jA} instead of \setarab\<جاء> \textit{jA'}  `he came'), a common phenomenon in many 
%modern 
Arabic dialects \cite{Habash:2010:introduction}. 
% Although sometimes it is represented by its support letter (\sethebrew\<רויא> \setarab\<رؤيا> \textit{r{\WHAMZA}yA}) or by lengthening its vocalization (\sethebrew\<סילת> \setarab\<سئلت> \textit{s{\YHAMZA}lt}).
Although sometimes it is represented by its support letter, e.g., \sethebrew\<רויא> \setarab\<رويا> \textit{rwyA} for \setarab\<رؤيا> \textit{r{\WHAMZA}yA} `vision' and \sethebrew\<סילת> \setarab\<سيلت> \textit{sylt} for \setarab\<سئلت> \textit{s{\YHAMZA}lt} `be asked'.

%Finally, following a transliteration approach is problematic when considering Aramaic or Hebrew words used in Judeo-Arabic. Labeling these words as either Arabic or Hebrew might not be the best approach. This distinction would be artificial, because they are also part of Judeo-Arabic and were treated in the same way as other words that came from Arabic. For example, some Hebrew root could be integrated into Arabic and then conjugated following its verbal system. This is the case of \sethebrew\<גיר>  \setarab\<غير>  \textit{{\GAYN}yr} ``convert to Judaism'' \cite{stillman2010judeo}.
Finally, transliteration is problematic for Hebrew words in JA. Labeling them as Arabic or Hebrew imposes an artificial distinction, since they function as part of JA and are treated like native Arabic words. For example, some Hebrew roots could be integrated into Arabic and then conjugated following its verbal system. This is the case of \sethebrew\<גיר>  \setarab\<غير>  \textit{{\GAYN}yr} `convert to Judaism' \cite{stillman2010judeo}. In Arabic texts derived from JA, such borrowings are typically translated rather than transliterated.

\subsection{Transliteration into Arabic Script}

\paragraph{Latin-based Arabic Transliteration}
Several studies have investigated the transliteration of Latin-based scripts into Arabic, with the most extensively studied case being \textit{Arabizi}, a non-standard romanization of Arabic widely used in informal online communication \cite{chalabi-gerges-2012-romanized,voss-etal-2014-finding,darwish-2014-arabizi,al-badrashiny-etal-2014-automatic,eskander-etal-2014-foreign,guellil:2017,younes:2018,shazal-etal-2020-unified,nagoudi-etal-2022-arat5}. While Arabizi has received the most attention when it comes to transliterating into Arabic, other Latin-script languages have also been explored. \newcite{micallef-etal-2023-exploring} treated Maltese as a dialect of Arabic, and investigated the effect of transliterating Maltese into Arabic script on downstream NLP tasks. Previous work has also addressed the problem of diacritizing and transliterating foreign words, particularly English proper nouns, into fully diacritized Arabic \cite{mubarak:2009:diacritization,darwish-etal-2017-arabic,bondok-etal-2025-proper}.

\paragraph{Judeo-Arabic to Arabic Transliteration} In the case of JA, \newcite{bar2015} proposed a three-step pipeline: identifying code-switched words to distinguish Hebrew words from Arabic, applying a character-level statistical MT model for transliteration, and using a recurrent neural network (RNN) for post-correcting a limited set of orthographic errors. \newcite{terner-etal-2020-transliteration} proposed an RNN to perform end-to-end transliteration. More recently, \newcite{mitelman-etal-2024-code} adopted a two-step approach: first identifying code-switched words in Hebrew, and then transliterating JA to Arabic script. Both of their components leverage HeArBERT \cite{rom2024training}, a BERT-based model pretrained on Arabic and Hebrew.
% In our work, we compare our results directly with this model.

While prior work on JA transliteration has made important contributions toward making JA accessible to Arabic readers, it suffers from several limitations in data preprocessing, evaluation metrics, and reproducibility. Data preprocessing pipelines vary considerably, particularly in how code-switched Hebrew words, orthographic markers (e.g., diacritics, hamzas, dots), and punctuation are handled. All previous studies rely on extracted parallel JA and Arabic texts aligned from published books; however, they adopt different alignment algorithms and uniformly remove Arabic translations corresponding to Hebrew citations (\S\ref{sec:ling-facts}) from the target Arabic data. Orthographic preprocessing also differs: \newcite{bar2015} retain diacritics, \newcite{terner-etal-2020-transliteration} and \newcite{mitelman-etal-2024-code} remove them; \newcite{mitelman-etal-2024-code} further remove hamzas and dots from the Hebrew input and retain Hebrew citations in the output, producing mixed-script transliterations. Punctuation is also inconsistently treated, with \newcite{bar2015} and \newcite{terner-etal-2020-transliteration} preserving it, while \newcite{mitelman-etal-2024-code}  discard it entirely. Evaluation metrics also vary: \newcite{bar2015} reported character-level accuracy, \newcite{terner-etal-2020-transliteration} used letter error rate, while \newcite{mitelman-etal-2024-code} adopted character-level precision, recall, F\textsubscript{1}, and accuracy. Finally, prior work uses different data sources for training and evaluation.

None of the previous studies make their models or data publicly available, with the exception of \newcite{mitelman-etal-2024-code}, who released their pretrained models and data. However, the released data is provided only at the character level without sentence boundaries or punctuation, limiting its utility for full-text evaluation. We therefore benchmark their system under our setup and treat it as the only available baseline for comparison. Comparisons with earlier studies remain infeasible due to the absence of released models and outputs.
% Due to substantial differences in data preprocessing, evaluation metrics, and the lack of comparable system outputs, direct comparison with prior work is not feasible.

\textbf{In our work}, as in prior studies, the goal is to facilitate access to JA texts for Arabic readers. However, our approach differs in several key aspects. We preserve the full input text structure, retaining Hebrew tokens and punctuation during processing, and restrict evaluation to Arabic tokens.
% while ignoring punctuation.
We systematically evaluate a range of models \textit{without} any training on JA data and explore systems ranging from simple character mappings to LLMs. Inspired by \newcite{bar2015}, we incorporate a post-correction stage to address orthographic errors; however, rather than relying on task-specific models targeting a limited set of orthographic errors, we leverage state-of-the-art pretrained Arabic grammatical error correction systems. Finally, we address an aspect overlooked in prior work by analyzing the impact of dotless versus dotted Hebrew orthography on transliteration performance, and demonstrate how transliteration benefits downstream Arabic NLP applications, such as morphosyntactic tagging and MT.

% We do not tackle code-switching
% Benchmarking LLMs on Jude-Arabic to Arabic transliteration
% Transliterate-then-correct using a two step approach
% We demonostrate the effectiveness of transliteration on downstream tasks: pos tagging, MT, etc.

\section{Data}
\label{sec:data}
% \begin{itemize}
%     \item Different data sources
%     \item Why Al-Khazari and not other source
%     \item Alignments and how they differ from previous work
%     \item Example of aligned sentence pairs
%     \item Stats (word-level, line level, segmentation)
% \end{itemize}

\subsection{Sources}
\label{sec:data-sources}
%Different data sources: a bunch of sources.
%Nabih.'s Arabic
%
%Why Al-Khazari and not other source
%No training... only test

We use the edition of \textit{Al-Khazari} by Nabih Bashir \cite{halevi2012kuzari}. This is the only edition of the text in Arabic and follows closely the JA original. However, because it was created with an Arabic audience in mind, the text was edited for clarity in some places and some of the terms from the JA original were replaced with their modern Arabic equivalents. This happened especially with words that come from Hebrew or with Biblical quotations that appeared in Hebrew in the original text. 
%For example, the word for poor in Hebrew (\sethebrew\<עני> \textit{{\AYN}ny}) is translated into its Arabic equivalent (\setarab\<فقير> \textit{fqyr}). 
For example, the Hebrew word \sethebrew\<עני> \textit{{\AYN}ny} `poor' is translated into its Arabic equivalent \setarab\<فقير> \textit{fqyr}. 
Thus, our Arabic source was not an exact human transliteration of the JA original, and we had to deal with the mismatches between the JA and the Arabic versions in an alignment step (\S\ref{sec:alignment}).

Nabih Bashir has produced other Arabic versions of JA texts:  the ``Introduction to the Mishnah'' by Maimonides and ``The book and beliefs and opinions'' by Saadia Gaon. Both were used by \newcite{mitelman-etal-2024-code} for training, although they only used Al-Khazari for evaluation. These two texts are unpublished and not publicly available, which is the primary reason we excluded them from our work and focused on Al-Khazari instead, though not the only one. 
% We believe that focusing on a single text might benefit our study.
% As discussed above (\S\ref{sec:ling-facts}), the use of the upper dot diacritic is not consistent across all JA texts, but it is consistent within a single text, and this allows us to properly understand the way it was used. Drawing conclusions about the upper dot diacritic based on its use across different texts, without considering context-specific usage, can give the misleading impression that it only introduces noise and should be ignored, as was done by \newcite{mitelman-etal-2024-code}. 
A further motivation, however, relates to the use of the upper dot diacritic. As discussed above (\S\ref{sec:ling-facts}), its usage is not consistent across all JA texts, but it is consistent within a single text. This consistency within Al-Khazari enables us to analyze the upper dot diacritic in a controlled manner and properly understand its function. In contrast, drawing conclusions from multiple texts without accounting for text-specific conventions can misleadingly suggest that the upper dot diacritic is mere noise, as assumed by \newcite{mitelman-etal-2024-code}.
Our findings show that the upper dot diacritic carries meaning, can aid disambiguation, and should therefore be preserved whenever it appears. Finally, our approach requires no training on JA data: it relies on deterministic transliteration followed by pretrained grammatical error correction models.

\subsection{Basic Character Mapping}
\label{sec:char-mapping}

%https://docs.google.com/spreadsheets/d/16NxF5OZp1X39ohCMXFykcdLwnfzfE_Bzt7XuqLBKNu4/edit?gid=0#gid=0
%%% %https://docs.google.com/spreadsheets/d/12Z0SnxtYQ1Tj3autsmwBDb5bFsywUNvirpY2EuD3I6g/edit?gid=0#gid=0
\begin{table}[ht!]
\centering
 \includegraphics[width=0.8\columnwidth]{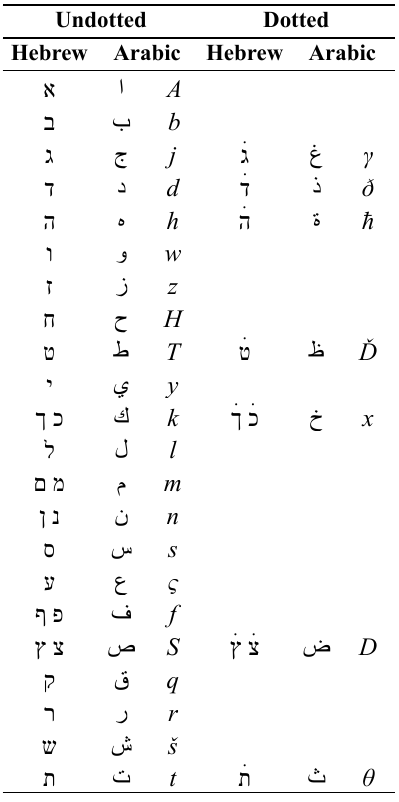}
    \caption{Arabic to Judeo-Arabic mapping used for transliteration. The dotted version takes into account the upper dot diacritic in the Judeo-Arabic text.\textsuperscript{\ref{translit}}}
\label{tab:map-characters}
\end{table}

For our basic one-to-one character mapping, we follow common conventions used in the field of JA studies \cite{blau1961grammar,judeoarabic2020orthographies}. See our full mapping in Table~\ref{tab:map-characters}. 
Our choices were later confirmed by the success rate of our transliteration. Our transliteration results using JA text with the upper dot diacritic (64.9\%) outperformed those without it (53.0\%) (\S\ref{sec:trans-results}). Examples where the upper dot diacritic made a difference include
{\sethebrew \<יכׄאטבה>} {\setarab \<يخاطبه>} \textit{yxATbh} `he speaks to him',~\phantom{\setarab \<ا>}
{\sethebrew\<מרצׄי>}
\setarab\<مرضي> \textit{mrDy} `satisfactory', and
\sethebrew\<תגׄיר>~\setarab\<تغير> \textit{t{\GAYN}yr} `he changed', whose respective dotless versions are \textit{incorrect}:
{\sethebrew \<יכאטבה>} {\setarab\<يكاطبه>} \textit{ykATbh},
\sethebrew\<מרצי> \setarab\<مرصي> \textit{mrSy}, and
\sethebrew\<תגיר> \setarab\<تجير> \textit{tjyr}.

\subsection{Preprocessing and Alignment}
\label{sec:alignment}
%mapping
%
%alignment
%
%validation.
%
%Labeling; Pnx, hebrew... etc.

Both the JA source and the Arabic reference transliteration produced by Nabih Bashir were obtained from Sefaria, an online library of Jewish texts.\footnote{\url{https://www.sefaria.org/Kuzari}} The files were divided into the same number of sections, which varied in length from a single sentence to several paragraphs. Footnotes in the Arabic reference were removed, after which we created two parallel text files with each section aligned to a line. We then reviewed sections with substantial word imbalances between the JA and Arabic texts and, after close reading, decided to drop 16 lines.\footnote{The lines dropped were: 130, 131, 203, 229, 230, 231, 232, 233, 234, 235, 242, 243, 291, 292, 303, 321.} 

% We first generated an Arabic transliteration from the original Judeo-Arabic (JA) text using the character mapping described above, preserving upper dot diacritics. 

% We then aligned the Arabic reference and the machine-generated transliteration at the word level using the method of \citet{khalifa-etal-2021-ced}. If a word in the machine-generated transliteration could not be aligned to any word in the Arabic reference, we inserted an \textit{UNK} token into the reference to preserve alignment structure. Conversely, unmatched words from the Arabic reference were discarded. Because the machine-generated transliteration maintains a one-to-one correspondence with the JA source, this alignment effectively maps JA tokens to their counterparts in the Arabic reference. Finally, we removed the upper dot diacritics from the JA source and used it to produce an alternative version of the machine-generated transliteration.

We first produced an automatic Arabic transliteration of the JA source using the character mapping described in \S\ref{sec:char-mapping}, accounting for the upper dot diacritic. Before alignment, Arabic diacritics were removed from the reference as commonly done in Arabic NLP \cite{elgamal-etal-2024-arabic,inoue2026diacritics}, and punctuation was separated in both the reference and the automatic transliteration.

We then aligned the Arabic reference and the automatic transliteration using word-level alignment \cite{khalifa-etal-2021-ced}. Words in the automatic transliteration with no match in the reference were mapped to an unknown token (\textit{UNK}) to preserve alignment, while unmatched insertions from the reference were discarded. Because the transliteration maintains a one-to-one correspondence with the JA source, this procedure effectively maps JA words to their Arabic counterparts. Finally, after completing the three-way alignment between the JA source, automatic transliteration, and reference, we removed upper dot diacritics from the JA source to generate an alternative \textit{dotless} Arabic transliteration.

% If a word from the Arabic automatically generated transliteration cannot be mapped to any word from the Arabic reference, we introduce an unknown (\textit{UNK}) word in the Arabic reference to preserve the alignment between both texts. Insertions from the Arabic reference that cannot be mapped to a word from the automatically generated Arabic transliteration are dropped. Because the generated transliteration maintains a one-to-one correspondence with the JA source, this alignment effectively maps JA words to their counterparts in the Arabic reference. After aligning the JA source, the automatically generated Arabic transliteration, and the Arabic reference (i.e., three-way alignment), we removed the upper dot diacritics from the JA source and generated an alternative Arabic transliteration.

For evaluation, each word in the JA source was labeled as JA, Hebrew, or punctuation. Hebrew words were identified primarily through the presence of biblical quotations; in the Arabic reference, such words are usually fully diacritized and translated, rather than transliterated (\S\ref{sec:data-sources}).

%To validate the labeling, we conducted human evaluation on a subset of 1,000 words and found only three errors. 
To validate the labeling, two independent annotators proficient in both Arabic and Hebrew evaluated a random subset of 1,000 words and reached full agreement, identifying only three labeling errors.
In all cases, a word labeled as JA was in fact of Hebrew origin and had been translated into Arabic in the reference: \phantom{\setarab \<س>}\sethebrew\<והמשכילים> \textit{whm{\SHIN}kylym} as \setarab\<والعقلاء> \textit{wAl{\AYN}qlA'} `and the wise ones'; \sethebrew\<יבינו>, \textit{ybynw} as \setarab\<يفهمونه> \textit{yfhmwnh} `they understand it'; and the JA abbreviation \sethebrew\<תעׄ> \textit{t{\AYN}} as \setarab \<تعالى> \textit{t{\AYN}Al{\AMAQ}} `Exalted'.

It is important to note that, both Hebrew words and punctuation were excluded from evaluation. Hebrew words were excluded because their Arabic counterparts are translations, not transliterations, and therefore not meaningful for transliteration evaluation. Punctuation was excluded because it is inconsistently used between the JA source and the Arabic reference; where it is often changed, added, or removed in the Arabic edition, making it unreliable as a point of comparison. See Appendix~\ref{app:align} Table~\ref{tab:alignment} for an alignment example.

% For evaluation, we labeled each word according to three categories: punctuation, Hebrew and Judeo-Arabic. Hebrew words were identified because in the Judeo-Arabic source they usually appeared in the context of biblical quotations. In those cases, they included Hebrew diacritics, so all the words that include Hebrew diacritics were labeled as Hebrew. It is important to note that Hebrew words and punctuation were excluded from the evaluation. In the first case, because Hebrew quotations were translated in the Arabic source, rather than transcribed. In the case of punctuation, because it is not relevant and inconsistent.

%, explained above in \S\ref{sec:ling-facts})

\subsection{Statistics}
\label{sec:data-stats}
%\begin{itemize}
%    \item how many paragraphs, sentences, and words?
%\end{itemize}
Our dataset is made of 325 paragraphs, 1,286 sentences and 46,529 words. Of those words, 4,576 (9.8\%) are labeled as punctuation, 916 (2\%) as Hebrew and 41,362 (88.3\%) as JA. Among the JA words, 856 were not aligned to anything in the Arabic reference, due to deletions or alignment errors. In total, these categories account for 6,348 words (13.6\%), which are excluded from transliteration evaluation. The evaluation is thus performed on the remaining 40,506 words (86.5\%).
% \textcolor{red}{How was the data labeled?}

\section{Approach}
\label{sec:approach}
Our goal is to transliterate JA texts written in Hebrew script into readable and fluent Arabic script. To this end, we propose a two-step approach. The first step starts with \textbf{transliteration} to convert the Hebrew script into Arabic. However, this step alone often results in Arabic text that is orthographically erroneous (\S\ref{sec:ling-facts}). We address this by introducing a second step of \textbf{post-correction}, which treats the output of transliteration as noisy Arabic and applies grammatical error correction (GEC) models to improve fluency and correctness.

\subsection{Transliteration}
\label{sec:transliteration}

\paragraph{Character Mapping}
We implement a rule-based character mapping approach to convert JA from Hebrew to Arabic script using manually curated mappings. These mappings are informed by linguistic analyses of JA orthography, as outlined in \S\ref{sec:ling-facts} and \S\ref{sec:char-mapping}. It is important to note that this type of direct transliteration often results in orthographic spelling errors and inconsistencies, such as missing hamzas, incorrect use of dots, and irregular morphological forms.

\paragraph{LLMs}
We evaluate OpenAI's GPT-3.5-turbo and GPT-4o \cite{gpt4}, prompting them to directly transliterate JA from Hebrew to Arabic script. The models are accessed via the OpenAI API and evaluated in a zero-shot setting. The prompts we use are in Table \ref{tab:prompts} in Appendix \ref{app:prompts}.

\subsection{Post-Correction}
\label{sec:correction}
While transliteration maps JA text from Hebrew to Arabic script, it does not guarantee orthographic or grammatical correctness. To address these challenges, we introduce a post-correction stage aimed at producing fluent, orthographically correct Arabic. We treat this as a Modern Standard Arabic (MSA) GEC task and explore both publicly available pretrained MSA GEC systems and LLMs. 

For pretrained systems, we use the sequence-to-sequence models developed by \newcite{alhafni-etal-2023-advancements}, including their best-performing vanilla model (Seq2Seq) as well as variants that incorporate additional signals such as morphological preprocessing and grammatical error detection (Seq2Seq+MG). We also evaluate the recently introduced text editing GEC system by \newcite{alhafni-habash-2025-enhancing}, which reformulates GEC as a sequence tagging problem (\textsc{sweet}).

In addition, we evaluate LLMs as general-purpose Arabic GEC systems. We experiment with GPT-3.5-turbo and GPT-4o, along with two open-source Arabic-centric models: Fanar \cite{fanar2025} and Jais-13B-Chat \cite{sengupta2023jais}. GPT-3.5-turbo, GPT-4o, and Fanar are accessed through the OpenAI API, while Jais is prompted using Hugging Face Transformers \cite{wolf-etal-2020-transformers}. Following the prompting strategy of \newcite{alhafni-habash-2025-enhancing}, we use English prompts in a zero-shot setting (Table~\ref{tab:prompts} Appendix~\ref{app:prompts}).

Details about all the models and the hyperparameters we use are provided in Appendix \ref{app:model-details}.

\section{Experimental Results}
\subsection{Evaluation Metrics}
\label{sec:metrics}
We use separate evaluation metrics for the transliteration and post-correction stages, aligned with the goals of each task.

For \textbf{transliteration}, We rely on the word-level alignment described in \S\ref{sec:alignment} and report exact match accuracy, which measures the percentage of transliterated JA words in Hebrew script that exactly match the gold Arabic reference. This metric reflects the system's ability to accurately map Hebrew words to Arabic script. Hebrew words and punctuation are excluded from the evaluation, as they are either fully translated or inconsistently used, and do not contribute meaningfully to the  task (\S\ref{sec:alignment}). 

For \textbf{post-correction}, we use the MaxMatch (M\textsuperscript{2}) scorer \cite{dahlmeier-ng-2012-better}, a standard metric for evaluating GEC systems. The M\textsuperscript{2} scorer computes overlap between the system's edits and the gold-standard edits derived from the target Arabic, reporting precision (P), recall (R), F\textsubscript{1}, and F\textsubscript{0.5} scores. The F\textsubscript{0.5} metric places twice the weight on precision relative to recall, emphasizing the correctness of the edits over their coverage. Gold edits are generated by aligning the automatic \textit{dotless} transliterated Arabic (\S\ref{sec:alignment}) with the Arabic reference using the algorithm proposed by \newcite{alhafni-etal-2023-advancements}. Unlike in transliteration, Hebrew words and punctuation are not excluded in GEC. This is because preserving the full sentence structure ensures that the model's corrections are evaluated in context, accurately reflecting its performance without ignoring potential errors these elements may introduce or correct.

% https://docs.google.com/spreadsheets/d/1BTcrwJKwmvIgHKZEOdnqmd4YM8rQSvx2iN9NE0EM-Fc/edit?gid=1638651373#gid=1638651373
\begin{table}[t!]
\setlength{\tabcolsep}{4pt}
\centering
\begin{tabular}{l cc}
\toprule
& {\bf Dotless} & {\bf Dotted} \\
\hline

\newcite{mitelman-etal-2024-code} & \textbf{70.0} & \textbf{70.1} \\\hline
CharMapper &  53.0 & 64.9  \\
GPT-3.5-turbo & 32.8 & 32.2  \\
GPT-4o &  52.9 & 38.6  \\
\bottomrule
\end{tabular}
\caption{Transliteration results in terms of word-level exact match accuracy. Dotted and Dotless refer to whether the input Judeo-Arabic text includes dots or not.}
%\textbf{M+~(2024)} refers to \newcite{mitelman-etal-2024-code}.}
\label{tab:trans-res}
\end{table}

\begin{table*}[ht!]
\centering
\begin{tabular}{l ccccc|ccccc}
\toprule
& \multicolumn{5}{c|}{\bf Dotless} & \multicolumn{5}{c}{\bf Dotted} \\
 & {\bf P} & {\bf R} & {\bf F\textsubscript{1}} & {\bf F\textsubscript{0.5}} & {\bf Acc.} & {\bf P} & {\bf R} & {\bf F\textsubscript{1}} & {\bf F\textsubscript{0.5}} & {\bf Acc.} \\
\hline

\newcite{mitelman-etal-2024-code} &   58.5 & 14.8 & 23.6 & 36.7 & 70.0 & 58.4 & 14.7 & 23.5 & 36.7 & 70.1  \\\hline
\hspace{1em}$\Rightarrow$ Seq2Seq &   75.0 & 26.6 & 39.3 & 55.0 & 82.9 & 74.9 & 26.7 & 39.3 & 55.0 & 82.8  \\
\hspace{1em}$\Rightarrow$ Seq2Seq+MG &   75.6 & 26.9 & 39.7 & 55.5 & 82.9 & 75.5 & 27.4 & 40.2 & 55.9 & 82.9  \\
\hspace{1em}$\Rightarrow$ \textsc{sweet} &   76.9 & 23.5 & 36.0 & 52.9 & 83.9 & 77.0 & 23.5 & 36.0 & 52.9 & 84.1  \\
\hline
\hspace{1em}$\Rightarrow$ GPT-3.5-turbo &   74.0 & 20.8 & 32.5 & 49.0 & 78.1 & 74.1 & 21.6 & 33.4 & 49.8 & 79.3  \\
\hspace{1em}$\Rightarrow$ GPT-4o &  77.9 & 23.8 & 36.4 & 53.5 & 86.1 & 77.7 & 23.5 & 36.1 & 53.2 & 87.4  \\
\hspace{1em}$\Rightarrow$ Fanar &  66.4 & 9.7 &  17.0 & 30.7 & 61.0 & 66.8 & 11.5 & 19.6 & 34.0 & 61.7  \\
\hspace{1em}$\Rightarrow$ Jais-13B-Chat &   54.0 & 3.8 &  7.0 &  14.7 & 24.9 & 53.6 & 4.2 &  7.8 &  16.0 & 24.7 \\\hline\hline

CharMapper &  \textbf{100} & 0 & 0 & 0 & 53.0 & 75.5 & 14.1 & 23.8 & 40.4  & 64.9 \\\hline
\hspace{1em}$\Rightarrow$ Seq2Seq &  72.8 & 27.2 & 39.6 & 54.5 & 74.5 & 81.2 & 30.3 & 44.1 & 60.8 & 85.2 \\
\hspace{1em}$\Rightarrow$ Seq2Seq+MG &  73.4 & \textbf{29.0} & 41.6 & 56.2 & 76.3 & 82.5 & 30.8 & 44.9 & 61.8 & 86.9 \\
\hspace{1em}$\Rightarrow$ \textsc{sweet} &  71.8 & 28.5 & 40.8 & 55.1 & 76.5 & 80.1 & \textbf{32.9} & \textbf{46.7} & 62.3 & 86.0 \\
\hline
\hspace{1em}$\Rightarrow$ GPT-3.5-turbo &  72.1 & 23.9 & 35.9 & 51.4 & 75.6 & 78.7 & 27.5 & 40.8 & 57.4 & 82.6 \\
\hspace{1em}$\Rightarrow$ GPT-4o &  82.0 & 28.4 & \textbf{42.2} & \textbf{59.5} & \textbf{86.4} & \textbf{85.9} & 30.7 & 45.2 & \textbf{63.1} & \textbf{90.4} \\
\hspace{1em}$\Rightarrow$ Fanar &  65.5 & 13.9 & 23.0 & 37.7 & 67.6 & 71.9 & 18.9 & 29.9 & 46.0 & 74.0 \\
\hspace{1em}$\Rightarrow$ Jais-13B-Chat &  48.0 & 3.4 & 6.4 & 13.3 & 31.4 & 53.2 & 4.2 & 7.8 & 16.0 & 31.1 \\

\bottomrule
\end{tabular}
\caption{Post-correction results for various systems on transliterated Arabic text derived from either dotless or dotted Judeo-Arabic Hebrew script.}
%M+ (2024) refers to \cite{mitelman-etal-2024-code}.}
\label{tab:gec-res}
\end{table*}

% https://docs.google.com/spreadsheets/d/1BTcrwJKwmvIgHKZEOdnqmd4YM8rQSvx2iN9NE0EM-Fc/edit?gid=1638651373#gid=1638651373
\begin{table*}[ht]
\setlength{\tabcolsep}{2.5pt}
\centering
\begin{tabular}{l cccccc|cccccc}
\toprule
& \multicolumn{6}{c|}{\bf Dotless} & \multicolumn{6}{c}{\bf Dotted} \\
 & \textbf{Match} & \textbf{Match'} & \textbf{POS} & \textbf{Diac} & \textbf{Lex} & \textbf{Tok} & \textbf{Match} & \textbf{Match'} & \textbf{POS} & \textbf{Diac} & \textbf{Lex} & \textbf{Tok} \\
\hline
GPT-3.5-turbo  &  32.8 & 35.8 & 46.3 & 33.4 & 38.0 & 35.7 & 32.2 & 35.6 & 46.7 & 33.1 & 37.7 & 35.5 \\
GPT-4o  &  52.9 & 53.5 & 59.3 & 51.9 & 55.1 & 53.4 & 38.6 & 39.2 & 44.2 & 37.9 & 40.6 & 39.2 \\\hline

\newcite{mitelman-etal-2024-code}  & 70.0 & 83.9 & 85.2 & 80.0 & 83.4 & 83.6 & 70.0 & 84.0 & 85.2 & 80.0 & 83.5 & 83.7\\
% \hspace{.8em}$\Rightarrow$ Seq2Seq+MG &  82.8 & 83.6 & 86.0 & 80.2 & 80.2 & 84.2 & 82.9 & 83.7 & 86.1 & 80.2 & 80.3 & 84.2\\
\hspace{.8em}$\Rightarrow$ GPT-4o &  86.1 & 86.4 & 89.1 & 83.5 & 87.0 & 86.3 & 87.4 & 87.7 & 90.3 & 84.7 & 88.2 & 87.5\\\hline

CharMapper &  53.0 & 71.2 & 76.9 & 67.8 & 71.4 & 71.0 & 64.9 & 85.8 & 88.5 & 82.7 & 86.0 & 85.5 \\
% \hspace{.8em}$\Rightarrow$ \textsc{sweet} &  76.5 & 77.2 & 83.5 & 74.6 & 77.8 & 77.1 & 86.0 & 87.2 & 90.5 & 85.0 & 87.9 & 87.0 \\
\hspace{.8em}$\Rightarrow$ GPT-4o & \textbf{86.4} & \textbf{86.7} & \textbf{92.4} & \textbf{84.7} & \textbf{87.7} & \textbf{86.6} & \textbf{90.4} & \textbf{90.6} & \textbf{93.9} & \textbf{88.9} & \textbf{91.5} & \textbf{90.6} \\

\bottomrule
\end{tabular}
\caption{Morphological tagging results in terms of word-level accuracy for Arabic outputs produced by various transliteration systems, using either dotted or dotless Judeo-Arabic input.}
%M+ (2024) refers to \newcite{mitelman-etal-2024-code}.}
\label{tab:morph-res}
\end{table*}

\subsection{Transliteration Results}
\label{sec:trans-results}
Table~\ref{tab:trans-res} presents transliteration results measured by word-level exact match accuracy. The model by \newcite{mitelman-etal-2024-code} achieves the highest overall accuracy on both dotted and dotless input. This is expected, as it was trained directly on JA data. Interestingly, its performance on dotted and dotless text is nearly identical, despite the model being trained only on dotless input.

The rule-based character mapping approach, henceforth referred to as \textit{CharMapper}, outperforms both GPT-3.5 and GPT-4o on dotted and dotless JA inputs, with particularly strong performance on dotted text. In contrast, the LLMs perform better on dotless input compared to dotted input, suggesting that they are more effective at handling dotless input.
%We speculate this may be the result of the low frequency of the upper dot diacritic, which may interact negatively with the LLM's tokenization.

\subsection{Transliteration \& Post-Correction Results}
Table~\ref{tab:gec-res} presents transliteration and post-correction results on transliterated Arabic derived from dotless and dotted JA input, using the baseline model of \newcite{mitelman-etal-2024-code} and our rule-based system \textit{CharMapper}.

Before post-correction, \newcite{mitelman-etal-2024-code} shows identical GEC performance on dotless and dotted input, mirroring its transliteration trend (\S\ref{sec:trans-results}). In contrast, \textit{CharMapper} records an F\textsubscript{0.5} of 0 on dotless input, as evaluation compares its raw transliteration to the gold reference (\S\ref{sec:metrics}). On dotted input, it outperforms \newcite{mitelman-etal-2024-code} in GEC (40.4 F\textsubscript{0.5}) but lags in transliteration accuracy.

Post-correction substantially improves GEC and transliteration for both \newcite{mitelman-etal-2024-code} and \textit{CharMapper}, except for Fanar and Jais on \newcite{mitelman-etal-2024-code} outputs and Jais on \textit{CharMapper}. For \newcite{mitelman-etal-2024-code}, dotless and dotted results remain similar, with Seq2Seq+MG giving the best GEC scores (55.5 dotless, 55.9 dotted F\textsubscript{0.5}) and GPT-4o the best transliteration accuracy (86.1 dotless, 87.4 dotted). For \textit{CharMapper}, dotted input yields consistently larger gains compared to dotless, with GPT-4o leading on both GEC (59.5 dotless, 63.1 dotted F\textsubscript{0.5}) and transliteration (86.4 dotless, 90.4 dotted).

These results highlight two key points: (1) post-correction consistently improves performance, and (2) despite its simplicity, \textit{CharMapper} with the upper dot diacritic, when combined with GEC, surpasses the JA-specific \newcite{mitelman-etal-2024-code} model under the same post-correction setting.

% Overall, post-correction performance is consistently higher when applied to transliterated Arabic outputs derived from dotted JA input compared to dotless input.

% Among the pretrained GEC systems, the Subword Edit Error Tagger (\textsc{sweet}) \cite{alhafni-habash-2025-enhancing} achieves the best performance on dotted text with an F\textsubscript{0.5} score of 62.3 (21.9 F\textsubscript{0.5} points improvement over \textit{CharMapper}), as well as the highest recall and F\textsubscript{1} overall. The Seq2Seq system enriched with morphological preprocessing and grammatical error detection (Seq2Seq+MG) \cite{alhafni-etal-2023-advancements} achieves the highest precision among the pretrained models and slightly surpasses \textsc{sweet} in word-level accuracy, showing a 22 point gain over \textit{CharMapper}. On dotless input, Seq2Seq+MG outperforms \textsc{sweet} in terms of F\textsubscript{0.5}, although its accuracy remains slightly lower.

% Among the LLMs, GPT-4o yields the best results. On dotted input, it achieves the highest F\textsubscript{0.5} score (63.1), the highest word-level accuracy (90.4, 25.5 point improvement over \textit{CharMapper}), and the best overall precision (85.9). GPT-4o also leads on dotless input, with an F\textsubscript{0.5} score of 59.5 and an accuracy of 86.4 (33.4 point improvements over \textit{CharMapper}), outperforming both non-LLM baselines and other LLMs across all metrics.

% These results highlight the importance of post-correction after transliteration, particularly in resolving common orthographic errors.

\subsection{Downstream Tasks Results}
Most Arabic NLP tools assume input in Arabic script, making transliteration a necessary step for processing JA texts. Converting JA from Hebrew to Arabic script enables these texts to be analyzed and processed using existing Arabic NLP systems. To demonstrate the utility of transliteration, we evaluate its impact on two downstream tasks: morphosyntactic tagging and machine translation.

\subsubsection{Morphological Tagging}
We use the contextualized Arabic morphosyntactic tagger \cite{inoue-etal-2022-morphosyntactic} from CAMeL Tools \cite{obeid-etal-2020-camel} to obtain morphological tags for both the transliterated JA text and its gold Arabic reference. Although no gold morphological annotations exist for our dataset, we treat the tags generated for the gold Arabic reference as \textit{silver annotations}, allowing us to compare the tagger's output on the transliterated text against a high-quality baseline.
%
% Table~\ref{tab:morph-res} presents morphological tagging results over the Arabic transliteration outputs.
%produced from transliterated Judeo-Arabic text. 
We evaluate six systems: GPT-3.5 and GPT-4o used directly for transliteration, and four systems based on the model by \newcite{mitelman-etal-2024-code} and our \textit{CharMapper}. Each is applied either alone or followed by GPT-4o post-correction, the setup with the best transliteration results.

We report word-level accuracy across multiple morphological features, including exact match (\textbf{Match}), minimal spelling correction 
%minimal analyzer-based spelling correction of Alif-Hamzas, Alif-Maqsura, and Ta-Marbuta
(\textbf{Match'}), part-of-speech (\textbf{POS}), fully diacritized form (\textbf{Diac}), lemma (\textbf{Lex}), and tokenization (\textbf{Tok}). It is worth noting that the \textbf{Match} score corresponds to the transliteration accuracy reported earlier (\S\ref{sec:trans-results}) and reflects whether the system's output exactly matches the gold Arabic word form.

Table~\ref{tab:morph-res} presents morphological tagging results over Arabic outputs produced from transliterated JA input. For both GPT-3.5 and GPT-4o, outputs derived from dotless input yield better tagging performance, with GPT-4o showing a particularly large improvement over its dotted counterpart. In contrast, both \newcite{mitelman-etal-2024-code} and \textit{CharMapper} outperform the LLMs across all tagging dimensions, regardless of input type, with higher accuracy on dotted text, a difference especially pronounced for \textit{CharMapper}.

Post-correction with GPT-4o yields substantial improvements across all features for both \newcite{mitelman-etal-2024-code} and \textit{CharMapper}, on both dotless and dotted input. The strongest results are achieved with \textit{CharMapper} on dotted text, where performance is consistently higher, highlighting the value of the upper-dot diacritic and the effectiveness of combining \textit{CharMapper} with high-quality error correction to produce well-formed Arabic.

\subsubsection{Machine Translation}
To assess the impact of transliterating JA to Arabic on MT to English, we first generate \textit{silver translations} by translating the Arabic gold references into English using GPT-4o. We then translate the Arabic outputs of six systems into English using GPT-4o: GPT-3.5, GPT-4o, \newcite{mitelman-etal-2024-code} (alone or with GPT-4o post-correction), and \textit{CharMapper} (alone or with GPT-4o post-correction). All translations are generated in a zero-shot setting with English prompts (Table~\ref{tab:prompts}, Appendix~\ref{app:prompts}).
Table~\ref{tab:mt-res} presents Arabic-to-English MT results. Across all systems, translations  from dotted input outperform those from dotless input, except \newcite{mitelman-etal-2024-code}.
%The highest BLEU scores are achieved when post-correction is performed using GPT-4o.
The highest BLEU scores are achieved with GPT-4o post-correction.

% https://docs.google.com/spreadsheets/d/1BTcrwJKwmvIgHKZEOdnqmd4YM8rQSvx2iN9NE0EM-Fc/edit?gid=1638651373#gid=1638651373
\begin{table}[t]
\setlength{\tabcolsep}{4pt}
\centering
\begin{tabular}{l cc}
\toprule
& {\bf Dotless} & {\bf Dotted} \\
\hline
GPT-3.5-turbo & 5.1  & 3.7 \\
GPT-4o  & 23.8  & 24.6 \\\hline

\newcite{mitelman-etal-2024-code} &  \textbf{26.1} & 24.6 \\
% \hspace{1em}$\Rightarrow$ Seq2Seq+MG & 18.9 & 20.9 \\
\hspace{1em}$\Rightarrow$ GPT-4o & 25.5 & 27.1 \\\hline

CharMapper &  20.2 & 24.1 \\
% \hspace{1em}$\Rightarrow$ \textsc{sweet} & 20.0 & 26.3 \\
\hspace{1em}$\Rightarrow$ GPT-4o & 24.1 & \textbf{28.3} \\
\bottomrule
\end{tabular}
\caption{Arabic-to-English machine translation results in terms of BLEU for outputs generated by different transliteration systems, using either dotted or dotless Judeo-Arabic input.}
%M+ (2024) refers to \newcite{mitelman-etal-2024-code}.}
\label{tab:mt-res}
\end{table}

\section{Error Analysis}
%https://docs.google.com/spreadsheets/d/11JEUYPB1WPri9lp9Mxod4w3qrqyA-ssP/edit?gid=1858785820#gid=1858785820
% 46529	Words
% 325	Paragraph
% 6023	Excluded words
% 40506	Included words

Our dataset contains 46,529 words organized into 325 paragraphs. We exclude 6,023 words from evaluation due to punctuation, Hebrew content, or alignment errors, leaving 40,506 words for analysis. We present four error analysis studies below.

\subsection{Analysis of Dotting Effect}
Of the 40,506 analyzed words, 83.4\% contain no dotted letters, while 16.6\% contain dotted letters. The \textit{CharMapper} system correctly maps 52.9\% of all words with no dots (63.5\% relative) and 11.9\% of all words with dots (71.8\% relative).
The 11.9\% last mentioned is the difference between \textit{CharMapper} baseline on Dotted vs. Dotless input.
The vast majority of errors for both dotted and undotted words stem from missing additional dots or from Arabic-only letters in the reference, such as hamza forms. For example, {\sethebrew <סנה>} ‘year’ lacks a dot on the final letter, yielding {\setarab <سنه>} \textit{snh} instead of the correct {\setarab <سنة>} \textit{sn\TAMARBUTA}.
Incorrect dot insertion is rare, occurring in only 0.04\% of dotted words, and typically involves letters that do not permit dotting. For instance, the dot in {\sethebrew <הםׄ>} ‘they’ is superfluous and is ignored, resulting in the correct output {\setarab <هم>} \textit{hm}.

% {\sethebrew \<אן>} {\setarab \<ان>} \textit{An} `if'  
% {\sethebrew \<מרצׄיה>} {\setarab \<مرضية>} \textit{mrḍyh} `satisfactory'  
% {\sethebrew \<ידׄ>} {\setarab \<يد>} \textit{yd} `hand'  
% {\sethebrew \<הםׄ>} {\setarab \<هم>} \textit{hm} `they'

\subsection{Recovery from Dotless Input in GPT-4o}

% 7	כ	82	572	11	585	1250	47.7%
% 6	ג	26	207	0	275	508	54.1%
% 5	צ	42	247	5	643	937	69.2%
% 4	ת	31	118	7	646	802	81.4%
% 3	ט	13	74	8	390	485	82.1%
% 2	ה	26	25	7	281	339	85.0%
% 1	ד	42	93	16	1793	1944	93.1%
%multi	45	168	17	81	311	31.5%
 
Comparing the \textit{CharMapper} $\Rightarrow$ GPT-4o transliteration performance in Table~\ref{tab:gec-res} on Dotted (90.4\%) vs Dotless (86.4\%) input reveals a 4.0\% absolute difference, which is substantially smaller than the 11.9\% gap observed for the \textit{CharMapper} baseline. This suggests that GPT-4o is largely able to recover from the absence of dots in the dotless experimental setting. 
To unpack this overall robustness, we examine performance separately on words with and without dotted letters (in the Dotted input setting) and the effect of removing the dots in the Dotless input setting. We further analyze recovery behavior at the letter level.

Performance on words without dots (83.4\% of all words) differs only marginally between Dotted and Dotless input settings (89.8\% vs. 89.2\%), whereas the gap for words containing dots (16.6\%) is substantial (93.4\% vs. 72.1\%). 
This confirms that GPT-4o successfully recovers the majority of cases that would otherwise require dotted input, although recovery effectiveness varies considerably across letters. We group the letters into three categories:

\begin{itemize}
    \item 
    \sethebrew<ד>-\setarab<د>-\textit{d}\,/\,\sethebrew<דׄ>-\setarab<ذ>-\textit{\DHA} (93.1\%), 
    \sethebrew<ה>-\setarab<ه>-\textit{h}\,/\,\sethebrew<הׄ>-\setarab<ة>-\textit{\TAMARBUTA} (85.0\%), and
    \sethebrew<ת>-\setarab<ت>-\textit{t}\,/\,\sethebrew<תׄ>-\setarab<ث>-\textit{\THA} (81.4\%), 
    which resemble errors made by Arabic native speakers, arising from dialectal variation 
    %with respect to MSA 
    as well as typographic errors \cite{attia-etal-2012-improved,Habash:2018:unified}.

    \item 
    \sethebrew<ט>-\setarab<ط>-\textit{T}\,/\,\sethebrew<טׄ>-\setarab<ظ>-\textit{\ZA} (82.1\%) and
    \sethebrew<צ>-\setarab<ص>-\textit{S}\,/\,\sethebrew<צׄ>-\setarab<ض>-\textit{D} (69.2\%), 
    which primarily resemble Arabic native speaker typographic errors only.

    \item 
    \sethebrew<ג>-\setarab<ج>-\textit{j}\,/\,\sethebrew<גׄ>-\setarab<غ>-\textit{\GAYN} (54.1\%) and
    \sethebrew<כ>-\setarab<ك>-\textit{k}\,/\,\sethebrew<כׄ>-\setarab<خ>-\textit{x} (47.7\%), 
    which differ substantially from typical Arabic native speaker errors and are therefore more challenging for MSA GEC systems. % to handle.
\end{itemize}

Words containing multiple dotted letters in the Dotted setting are the most difficult to recover in the Dotless setting, with an accuracy of only 31.5\%.

\subsection{Analysis of Best System Errors}
The best-performing system (\textit{Dotted} $\Rightarrow$ \textit{CharMapper} $\Rightarrow$ GPT-4o) produces 3,896 errors (9.6\%). We examined the first 100 errors, occurring within the first 2,690 words.
% The total number of instances in our dataset are 46,853. From that total, 6,342 are excluded from evaluation because they are labeled as punctuation, Hebrew or the Arabic reference does not map to anything.  The subset for evaluation is 40,511 instances long. Our best-performing system (\textit{CharMapper} $\Rightarrow$ GPT-4o) produces 3,903 errors (9.63\%). We evaluated the first one hundred errors, which occur within the first 2,740 words.
% Table \ref{tab:error-ana} in Appendix \ref{app:errors} presents error categories examples which display the Hebrew source, the \textit{CharMapper} transliteration, the output after post-correction with GPT-4o, and the gold Arabic reference.
We grouped errors into six categories, summarized in Table~\ref{tab:error-ana} (Appendix~\ref{app:errors}).

\textbf{Valid Paraphrase} (36\%) covers cases where the Arabic gold reference is not a direct transliteration of the JA source but a close substitution. For example, the system output \setarab<قال> \textit{qAl} `he said' is evaluated against the reference \setarab<فقال> \textit{fqAl} `then he said'. \textbf{Unnecessary Change} (21\%) refers to cases in which the \textit{CharMapper} output is changed by GPT-4o but the change was not necessary.
% In the example, the expected output was \setarab<خلقة>  \textit{xlq\TAMARBUTA} but GPT-4o removed the final character \setarab\<خلق>  \textit{xlq}.
\textbf{Alif-Hamza} (18\%) refers to errors caused by a wrong transliteration and placing of the Hamza (e.g., \setarab<الآب> \textit{Al{\AMADDA}b} instead of \setarab\<الأب> \textit{Al{\AHAMZAUP}b} `the father'). \textbf{Wrong Word} denotes cases where the system output diverges entirely from the gold reference. For example, the system produces \setarab<وإنجاء> \textit{w{\AHAMZADN}njA'}  instead of the reference \setarab<وتسليم> \textit{wtslym} `salvation/deliverance'.
\textbf{Source Error} (7\%) refers to spelling mistakes in the JA source itself. For example, the source \sethebrew<אנה> \setarab<انه> \textit{Anh} should have been \sethebrew<אנא> \setarab<أنا> \textit{{\AHAMZAUP}nA} `I'.
% \textbf{Heb to Ara Translation} (2\%) refers to cases where the JA source is a Hebrew word and the Arabic reference provides its translation rather than a transliteration. For example, \setarab<والعقلاء> \textit{wAl{\AYN}qlA'} is the Arabic translation of the Hebrew \sethebrew<והמשכילים> \textit{whm{\SHIN}kylym}, both meaning `and the wise'. 
Errors that are not covered by the previous cases were categorized as \textbf{Other} (6\%).  

\subsection{Analysis of Hebrew Input}
Although Hebrew input words (2\% of the dataset) were excluded from transliteration evaluation, we manually analyzed a random sample of 100 cases to assess how our best-performing system handles them. Appendix \ref{app:heb-analysis} Table~\ref{tab:error-ana-heb} presents the categories. In 84\% of cases, the system did not match the reference: 46\% were due to Hebrew words being translated into Arabic in the reference, 17\% involved substitution with a different word, and 21\% fell into other categories such as deletions or insertions. Notably, in 16\% of cases the system produced the correct output, which is plausible given the shared cognates between Hebrew and Arabic.

\section{Conclusions and Future Work}
We presented a two-step approach for transliterating Judeo-Arabic texts from Hebrew to Arabic script, combining character-level mapping with post-correction to resolve orthographic and grammatical inconsistencies. Our results show that post-correction substantially improves transliteration quality and, despite requiring no training on Judeo-Arabic data, our approach outperforms previous work. We also benchmarked LLMs on this task for the first time and demonstrated the downstream benefits of transliteration on Arabic morphosyntactic tagging and Arabic-English machine translation. Our goal is to make Judeo-Arabic texts accessible to Arabic readers and compatible with modern NLP tools, and we will make our code and data publicly available to support future research.
% To support reproducibility and further research, we will release our code and data publicly.

In future work, we aim to address the challenge of code-switching by automatically detecting and processing Hebrew segments embedded in Judeo-Arabic text.
Specifically, we plan to develop  models that combine transliteration with selective translation, converting non-Arabic segments into Arabic script.
%Specifically, we plan to develop integrated models that combine transliteration with selective translation, converting non-Arabic segments into Arabic script while preserving the overall structure of the text.
Additionally, we intend to investigate alternative modeling strategies to improve transliteration accuracy and robustness across diverse Judeo-Arabic texts. We also plan to assess the utility of transliteration on other downstream Arabic NLP tasks, such as named entity recognition.%, and to develop user-friendly online tools for broader access.

\section*{Acknowledgement}
% Juabil cluster
We acknowledge the support of the High Performance Computing Center at New York University Abu Dhabi.
We thank the anonymous reviewers for their insightful and constructive comments.

\section*{Ethical Considerations}
We use publicly available datasets and language models.
We do not anticipate any potential risks associated with this work, as it does not involve the collection of personal data, sensitive content, or human subjects. 
We used AI writing assistance within the scope of ``Assistance
purely with the language of the paper'' described in the ACL Policy on Publication Ethics.

\section*{Limitations}
While our work demonstrates promising results, several limitations should be noted that may affect its broader applicability and reproducibility. First, our experiments rely on closed-source commercial LLMs, which are subject to periodic updates that are not publicly documented. This introduces some uncertainty and may affect the reproducibility of our results over time. Second, although our transliteration and post-correction pipeline handles Judeo-Arabic well at the sentence level, it does not explicitly address Hebrew segments. These code-switched elements are common in Judeo-Arabic texts and present unique challenges, particularly when they are morphologically integrated into the surrounding Arabic. Finally, our downstream evaluations rely on silver annotations: morphological tags from pretrained Arabic taggers and English translations generated by GPT-4o. While informative, these approximations are not substitutes for human-validated gold data and may introduce evaluation noise.

\bibliography{custom,camel-bib-v3,anthology-1,anthology-2}

% \onecolumn
%\newpage
\appendix
% \section{Character Mapping}
% \label{app:char-map}
% \noindent
% \begin{figure}[h!]
% \centering
%  \includegraphics[width=0.8\columnwidth]{JA-Ara-Map.pdf}
%     \caption{Arabic to Judeo-Arabic mapping used for transliteration. The dotted version takes into account the upper dot diacritic in the Judeo-Arabic text.}
% \label{tab:map-characters}
% \end{figure}

% \newpage
\section{Model Details}
\label{app:model-details}
\paragraph{Pretrained GEC Systems} The Seq2Seq and Seq2Seq+MG models introduced by \newcite{alhafni-etal-2023-advancements} are built on AraBART \cite{kamal-eddine-etal-2022-arabart} and publicly available at \url{https://github.com/CAMeL-Lab/arabic-gec}
. The Seq2Seq+MG variant incorporates morphological features and grammatical error detection signals obtained from external models, as described in \newcite{alhafni-etal-2023-advancements}. Seq2Seq and Seq2Seq+MG consists of 139M and 502M parameters, respectively.
The text editing model \textsc{sweet} \cite{alhafni-habash-2025-enhancing}, built on AraBERTv02 \cite{antoun-etal-2020-arabert}, has 135M parameters and is available at \url{https://github.com/CAMeL-Lab/text-editing}. For all pretrained GEC models, we use their respective default hyperparameters. Inference was done on a single A100 GPU.

\paragraph{LLMs} Jais-13B-Chat has 13B parameters, and Fanar has 8.7B. Architectural details for GPT-3.5-turbo and GPT-4o are unavailable, as they are not publicly released. GPT-3.5-turbo, GPT-4o, and Fanar were prompted using default hyperparameters, while Jais was run on an A100 GPU with temperature 0.3, nucleus sampling 0.9, max length 2048, and repetition penalty 1.2.

\onecolumn
\clearpage
\section{Alignment Example}
\label{app:align}

\begin{table*}[h]
\setlength{\tabcolsep}{6pt}
\centering
\small
\begin{tabular}{lccll}
\toprule
\textbf{JA} & \textbf{Transliteration} & \textbf{Reference} & \textbf{Label} & \textbf{Gloss} \\
\midrule
\RL{\sethebrew{פיציר}}    &  \cellcolor{green!5}\RL{\setarab{فيصير}} \textit{fySyr}  & \cellcolor{green!5}\RL{\setarab{فيصير}} \textit{fySyr}  & JA & `so he is' \\
\RL{\sethebrew{ענדנא}}    & \cellcolor{green!5}\RL{\setarab{عندنا}} \textit{{\AYN}ndnA} & \cellcolor{green!5}\RL{\setarab{عندنا}} \textit{{\AYN}ndnA} & JA & `to us' \\
\RL{\sethebrew{מרה}}      & \cellcolor{red!5}\RL{\setarab{مره}} \textit{mrh}   & \cellcolor{red!5}\RL{\setarab{مرة}}  \textit{mr{\TAMARBUTA}}  & JA & `sometimes' \\
\RL{\sethebrew{אֵל}}       & \cellcolor{yellow!5}\RL{\setarab{ال}}   \textit{Al}   & \cellcolor{yellow!5}\RL{\setarab{إله}} \textit{{\AHAMZADN}lh}   & Heb   & `God' \\
\RL{\sethebrew{רַחוּם}}    & \cellcolor{yellow!5}\RL{\setarab{رحوم}} \textit{rHwm}   & \cellcolor{yellow!5}\RL{\setarab{رحيم}}  \textit{rHym}  & Heb   & `compassionate' \\
\RL{\sethebrew{וְחַנּוּן}} & \cellcolor{yellow!5}\RL{\setarab{وحنون}}  \textit{wHnwn} & \cellcolor{yellow!5}\RL{\setarab{ورءوف}} \textit{wr'wf} & Heb   & `and graceful' \\
\RL{\sethebrew{ומרה}}      & \cellcolor{red!5}\RL{\setarab{ومره}} \textit{wmrh}    & \cellcolor{red!5}\RL{\setarab{ومرة}}  \textit{wmr{\TAMARBUTA}}   & JA & `other times' \\
\RL{\sethebrew{אֵל}}       & \cellcolor{yellow!5}\RL{\setarab{ال}}   \textit{Al}    & \cellcolor{yellow!5}\RL{\setarab{إله}}  \textit{{\AHAMZADN}lh}   & Heb   & `God' \\
\RL{\sethebrew{קַנֹּא}}    & \cellcolor{yellow!5}\RL{\setarab{قنا}}   \textit{qnA}  & \cellcolor{yellow!5}UNK                   & Heb   & `jealous' \\
\RL{\sethebrew{וְנוֹקֵם}}  & \cellcolor{yellow!5}\RL{\setarab{ونوقم}}  \textit{wnwqm} & \cellcolor{yellow!5}\RL{\setarab{ومنتقم}}  \textit{wmntqm} & Heb   & `and vengeful' \\
,                          & \cellcolor{yellow!5},                      & \cellcolor{yellow!5}UNK                   & Punc  & \\
\bottomrule
\end{tabular}
\caption{Alignment example. Green indicates successful automatic transliteration, red marks transliteration errors, and yellow highlights items excluded from evaluation (Hebrew and punctuation).}
\label{tab:alignment}
\end{table*}

 \section{Prompts}
\label{app:prompts}
\begin{table*}[h!]
\centering
\small
\begin{tabular}{@{}p{3.2cm}p{12cm}@{}}
\toprule
\textbf{Task} & \textbf{Prompt} \\
\midrule
Transliteration & 
\texttt{You are a transliteration system that can transliterate Judeo-Arabic text to Arabic. Please transliterate the following Judeo-Arabic sentence to Arabic without providing any explanation. The output should be in Arabic script.} \\
\addlinespace[0.5em]
GEC & 
\texttt{Please identify and correct any grammatical and spelling errors in the following sentence marked with the tag $<$input$>$ SRC $<$/input$>$. Make the minimal changes necessary to correct the sentence. Do not rephrase any parts of the sentence that are already grammatically correct, and avoid altering the meaning by adding or removing information. After making the corrections, output the revised sentence directly without providing any explanations. Remember to format the corrected output with the tag $<$output$>$ Your Corrected Version $<$/output$>$.} \\
\addlinespace[0.5em]
Machine Translation & 
\texttt{You are a machine translation system that can translate Arabic to English. Please translate the following Arabic sentence to English without providing any explanation.} \\
\bottomrule
\end{tabular}
\caption{Prompts used for each task in our experiments: transliteration, grammatical error correction (GEC), and machine translation.}
\label{tab:prompts}
\end{table*}
\clearpage

\section{Error Analysis}
\label{app:errors}

\begin{table*}[th]
\setlength{\tabcolsep}{3pt}
\small
\centering
\begin{tabular}{l c c c c c | c}
\toprule
\textbf{Error Type} & \textbf{\%} & \textbf{JA} & \textbf{CharMapper} & \textbf{CharMapper $\Rightarrow$ GPT-4o} & \textbf{Reference} & \textbf{Gloss} \\
\midrule
Valid Paraphrase        & 38 & \RL{\sethebrew{קאל}}       & \RL{\setarab{قال}} \textit{qAl}       & \RL{\setarab{قال}} \textit{qAl}       & \RL{\setarab{فقال}}     \textit{fqAl}   & `said' \\
Unnecessary Change      & 21 & \RL{\sethebrew{כׄלקהׄ}}     & \RL{\setarab{خلقة}} \textit{xlq{\TAMAR}}     & \RL{\setarab{خلق}}  \textit{xlq}     & \RL{\setarab{خلقة}}   \textit{xlq{\TAMAR}}    & `creation' \\

Alif-Hamza Error   & 18 & \RL{\sethebrew{אלאב}}      & \RL{\setarab{الاب}} \textit{AlAb}     & \RL{\setarab{الآب}}  \textit{Al{\AMADDA}b}    &  \RL{\setarab{الأب}}   \textit{Al{\AHAMZAUP}b}    & `the father' \\

Wrong Word              & 10 & \RL{\sethebrew{ותסלים}}    & \RL{\setarab{وتسليم}}  \textit{wtslym}  & \RL{\setarab{وإنجاء}}  \textit{w{\AHAMZADN}njA'}  & \RL{\setarab{وتسليم}}  wtslym\textit{}   & `salvation' \\

Source Error            & 7  & \RL{\sethebrew{אנה}}       & \RL{\setarab{انه}}  \textit{Anh}     & \RL{\setarab{إنه}}    \textit{{\AHAMZADN}nh}  & \RL{\setarab{أنا}} \textit{{\AHAMZAUP}nA} & `I'\\

% Heb to Ara Translation  & 2  & \RL{\sethebrew{והמשכילים}} & \RL{\setarab{وهمشكيليم}} \textit{whm{\SHIN}kylym} & \RL{\setarab{وهمشكيليم}} \textit{whm{\SHIN}kylym} & \RL{\setarab{والعقلاء}} \textit{wAl{\AYN}qlA'}  & And the wise \\

Other              & 6  & \RL{\sethebrew{מאיה}}       & \RL{\setarab{مايه}} \textit{mAyh}      & \RL{\setarab{خمسمائة}} \textit{xmsmA{\YHAMZA}{\TAMAR}}  & \RL{\setarab{ماية}}     \textit{mAy{\TAMAR}}   & `hundred' \\
\bottomrule
\end{tabular}
\caption{Error categories and examples.}
\label{tab:error-ana}
\end{table*}

 \section{Analysis of Hebrew Input}
\label{app:heb-analysis}

% \begin{table*}[th]
% \setlength{\tabcolsep}{5pt}
% \small
% \centering
% \begin{tabular}{l c c c c c | c}
% \toprule
% \textbf{Category} & \textbf{\%} & \textbf{JA} & \textbf{CharMapper} & \textbf{CharMapper $\Rightarrow$ GPT-4o} & \textbf{Reference} & \textbf{Gloss} \\
% \midrule

% Translation      & 46 & \RL{\sethebrew{וַיֹּאמַר}} & \RL{\setarab{ويامر}} \textit{wyAmr} & \RL{\setarab{ويأمر}} \textit{wy{\AHAMZAUP}mr} & \RL{\setarab{وقال}} \textit{wqAl} & And he said \\
% Substitution & 17 & \RL{\sethebrew{מִשְׁפְּחוֹת}} & \RL{\setarab{مشفحوت}} \textit{m{\SHIN}fHwt} & \RL{\setarab{مشفحوته}} \textit{m{\SHIN}fHwth} & \RL{\setarab{فقط}}  \textit{fqT}    & Only  \\
% % Deletion      & 17 & \RL{\sethebrew{אֲשֶׁר}}    & \RL{\setarab{اشر}} \textit{A{\SHIN}r}       & \RL{\setarab{الذي}} \textit{Alðy}        &                   & Which \\

% Other & 21 & \RL{\sethebrew{אֲשֶׁר}}    & \RL{\setarab{اشر}} \textit{A{\SHIN}r}       & \RL{\setarab{الذي}} \textit{Al{\THA}y}        &                   & Which \\\hline

% Correct       & 16 & \RL{\sethebrew{הוּא}}        & \RL{\setarab{هوا}} \textit{hwA}   & \RL{\setarab{هو}} \textit{hw}     & \RL{\setarab{هو}} \textit{hw} & He\\
% % Other & 21 & \RL{\sethebrew{וְאוֹכְלָה}} & \RL{\setarab{واوكله}} \textit{wAwklh}   & \RL{\setarab{وأكله}} \textit{wÂklh}   & \RL{\setarab{وأوخله}} \textit{wÂwxlh} & He ate  \\	

% \bottomrule
% \end{tabular}
% \caption{Hebrew input analysis.}
% \label{tab:error-ana-heb}
% \end{table*}

%%%%%%%%%%%

\begin{table*}[h!]
\setlength{\tabcolsep}{2pt}
\small
\centering
\begin{tabular}{l c   c c   c c c   c  c}
\toprule
                  &             &             &          \textbf{Hebrew}     &                     & \textbf{CharMapper} &        \textbf{Arabic} &                       &     \textbf{Arabic}    \\
\textbf{Category} & \textbf{\%} & \textbf{JA} & \textbf{Gloss} & \textbf{CharMapper} & \textbf{$\Rightarrow$ GPT-4o} & \textbf{Gloss}  & \textbf{Reference} & \textbf{Gloss} \\

\midrule

False Cognate      & 46 & \RL{\sethebrew{וַיֹּאמַר}} & `and he said' & \RL{\setarab{ويامر}} \textit{wyAmr} & \RL{\setarab{ويأمر}} \textit{wy{\AHAMZAUP}mr} & `and he orders' & \RL{\setarab{وقال}} \textit{wqAl} & `and he said' \\
Transliteration & 17 & \RL{\sethebrew{מִשְׁפְּחוֹת}} & `families' & \RL{\setarab{مشفحوت}} \textit{m{\SHIN}fHwt} & \RL{\setarab{مشفحوته}} \textit{m{\SHIN}fHwth} & [not Arabic]  &\RL{\setarab{فقط}}  \textit{fqT}    & `only'  \\
No Reference & 21 & \RL{\sethebrew{אֲשֶׁר}} & `which'   & \RL{\setarab{اشر}} \textit{A{\SHIN}r}       & \RL{\setarab{الذي}} \textit{Al{\THA}y}        &   `which' &              $\phi$  & $\phi$ \\\midrule

Correct       & 16 & \RL{\sethebrew{הוּא}}    & `he'    & \RL{\setarab{هوا}} \textit{hwA}    & \RL{\setarab{هو}} \textit{hw}  & `he'   & \RL{\setarab{هو}} \textit{hw} & `he'\\

\bottomrule
\end{tabular}
\caption{Analysis of different Hebrew input processing results.}
\label{tab:error-ana-heb}
\end{table*}

\end{document}